%% file: root.tex
\documentclass[letterpaper, 10 pt, conference]{ieeeconf}  

\IEEEoverridecommandlockouts                              

\usepackage{amsmath} 
\usepackage{amssymb}  
\usepackage{algorithmic}
\usepackage{algorithm}
\usepackage{array}
\usepackage[caption=false,font=normalsize,labelfont=sf,textfont=sf]{subfig}
\usepackage{xspace}
\usepackage{textcomp}
\usepackage{booktabs}   
\usepackage{xurl}       
\usepackage{makecell}
\usepackage[hidelinks]{hyperref}
\usepackage{array}
\usepackage{stfloats}
\usepackage{url}
\usepackage{verbatim}
\usepackage{graphicx}
\usepackage{cite}
\usepackage{cuted}   
\usepackage{capt-of} 
\usepackage{flushend}
\usepackage{xcolor}
\usepackage{tabularx}

\newcommand{\projectpage}{\href{https://wenzewwz123.github.io/Agentic-Loop/}{\texttt{https://wenzewwz123.github.io/Agentic-Loop/}}\xspace}

\title{\LARGE \bf
A Physical Agentic Loop for Language-Guided Grasping with Execution-State Monitoring  
}

\author{Wenze Wang, Mehdi Hosseinzadeh, Feras Dayoub
\\
\projectpage
\thanks{Authors are with the Australian Institute for Machine Learning (AIML), Adelaide University, Australia.}
}

\begin{document}

\maketitle
\thispagestyle{empty}
\pagestyle{empty}


\input{figures/teaser}


\begin{abstract}
Robotic manipulation systems that follow language instructions often execute grasp primitives in a largely single-shot manner: a model proposes an action, the robot executes it, and failures such as empty grasps, slips, stalls, timeouts, or semantically wrong grasps are not surfaced to the decision layer in a structured way. Inspired by agentic loops in digital tool-using agents, we reformulate language-guided grasping as a bounded embodied agent operating over grounded execution states, where physical actions expose an explicit tool-state stream. We introduce a physical agentic loop that wraps an unmodified learned manipulation primitive (grasp-and-lift) with (i) an event-based interface and (ii) an execution monitoring layer, \emph{Watchdog}, which converts noisy gripper telemetry into discrete outcome labels using contact-aware fusion and temporal stabilization. These outcome events, optionally combined with post-grasp semantic verification, are consumed by a deterministic bounded policy that finalizes, retries, or escalates to the user for clarification, guaranteeing finite termination. We validate the resulting loop on a mobile manipulator with an eye-in-hand D405 camera, keeping the underlying grasp model unchanged and evaluating representative scenarios involving visual ambiguity, distractors, and induced execution failures. Results show that explicit execution-state monitoring and bounded recovery enable more robust and interpretable behavior than open-loop execution, while adding minimal architectural overhead. For the source code and demo refer to our project page: \projectpage.
\end{abstract}


\input{sections/introduction_v2}
\input{sections/related_v2}
\input{sections/method_v2}
\input{sections/experiments_v2}

%
\input{sections/conclusion}
%








\bibliographystyle{IEEEtran}
\bibliography{references}

\end{document}

%% file: figures/teaser.tex
\begin{strip}
    \vspace*{-3.8\baselineskip} 
    \centering
    \includegraphics[width=\linewidth]{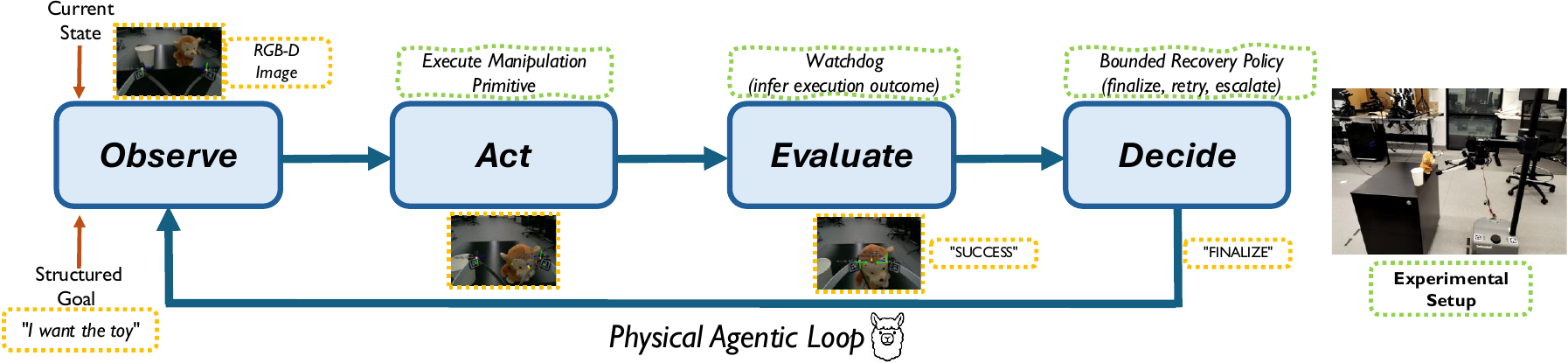}
    \vspace*{-1.7\baselineskip} 
    \captionof{figure}{\textbf{Physical agentic loop for language-guided grasping.} 
    In \emph{Observe}, the agent receives a structured goal (e.g., ``I want the toy''), and the current state (e.g., RGB-D observations). 
    In \emph{Act}, it executes an unmodified pre-trained manipulation primitive. 
    In \emph{Evaluate}, \emph{Watchdog} infers a discrete execution outcome (e.g., \texttt{SUCCESS}) from execution feedback and emits an outcome event. 
    In \emph{Decide}, a bounded recovery policy maps the outcome to an action (finalize, retry, or escalate/clarify), producing a decision event (e.g., \texttt{FINALIZE}) that closes the loop.
    Example snapshots alongside each stage illustrate one representative grasping sequence under the real-robot experimental setup.
    }
    \label{fig:teaser}
  \vspace*{-0.5\baselineskip} 
\end{strip}

%% file: sections/introduction_v2.tex
\section{Introduction}
\label{sec:intro}

Recent progress in digital tool-using agents (e.g., OpenClaw~\cite{openclaw}) has highlighted the value of the \emph{agentic loop}: systems that invoke tools, observe explicit execution states, and apply bounded retries or escalation when tools fail. In this setting, tool calls are first-class actions with a standardized lifecycle (e.g., start/progress/success/failure), which makes outcomes legible to the decision layer and enables principled recovery. Robotic manipulation, despite rapid advances in vision-language control and generalist policies, often lacks an analogous abstraction. Many systems still execute physical actions in a largely single-shot manner—plan once, execute once, and implicitly assume the world cooperates.

This gap is especially consequential for open-vocabulary, human-facing grasping, where the robot must act under partial observability, noisy perception, and unmodeled contact dynamics. The same instruction can yield empty grasps, weak grasps that slip during lift, stalls, timeouts, or grasps that are physically successful but semantically wrong (e.g., picking the wrong cup). Such failures are common in real deployments, yet they are rarely surfaced as structured signals that can inform downstream decisions. As a result, higher-level components may hallucinate success, over-trust perception confidence, or rely on brittle heuristics that do not generalize across objects, scenes, and contact conditions.

We argue that closing this gap requires importing the agentic-loop discipline into the physical domain, not by retraining the grasp policy but by making execution \emph{observable, interpretable, and actionable} to the agent. We do so by introducing a \textbf{physical agentic loop} for language-conditioned grasping (Figure~\ref{fig:teaser}). The core idea is to convert noisy, continuous execution evidence, such as gripper effort/current dynamics, gripper closure behavior, and post-action visual cues, into discrete, semantically meaningful execution states analogous to software-agent tool states. Concretely, we introduce \textbf{Watchdog}, a lightweight, low-latency outcome monitoring layer that infers grasp outcomes and failure modes (e.g., empty, weak, slip, stall, timeout) from gripper dynamics and emits structured outcome events with confidence. These events can be fused with a \emph{post-grasp vision verifier} that checks whether the grasped object matches the user’s intended target, addressing a critical failure class in language-guided grasping: executions that are physically successful but instruction-inconsistent. A dialogue and decision module then consumes this fused evidence and implements a \emph{bounded recovery policy}: it retries once when evidence indicates a recoverable physical failure, requests clarification when semantic evidence is ambiguous, and terminates safely when repeated action is unlikely to help. The resulting loop is interpretable and safety-bounded (limited retries, explicit escalation, and termination conditions), mirroring the reliability principles that make software agent loops robust.

This framing yields two practical advantages. First, it enables a \emph{wrapper-style} improvement to existing manipulation stacks, such as visual-force goal prediction and servo controllers~\cite{collins2024forcesight}, without modifying or retraining their models. Under this view, the underlying grasping components are treated as callable primitives whose outcomes are summarized through explicit execution-state events. Second, it supports rigorous, component-wise evaluation: by toggling outcome monitoring, semantic verification, and recovery behavior while holding the underlying controller fixed, we can isolate how each loop element contributes to robustness. In real-robot experiments on a Stretch mobile manipulator across scenarios that induce both physical failures and semantic confusions, we show that explicit execution-state monitoring combined with bounded recovery improves task success with minimal additional latency. More broadly, these results suggest that robust language-guided grasping often depends not only on stronger policies or perception models, but also on a reliable agentic loop that makes execution failures observable, interpretable, and recoverable at runtime.

\textbf{Contributions}. This paper makes three contributions:
\begin{itemize}
    \item We identify execution-time failure modes in language-guided grasping that cannot be reliably resolved from language or pre-grasp perception alone, and represent them as explicit execution states that the agent can use to decide whether to finalize, retry, or ask for clarification.
    \item We introduce \textbf{Watchdog}, a modular execution-outcome inference layer that maps gripper telemetry to discrete outcome events with confidence under real-time constraints. 
    \item We demonstrate a physical agentic loop that combines execution-state monitoring, bounded retry logic, and optional semantic conditioning to improve robustness on a real robot without retraining the underlying visual-force grasping model.
\end{itemize}

%% file: sections/related_v2.tex
\section{Related Work}
\label{sec:related}

\noindent \textbf{Agentic loops in digital tool-using agents.}
Software-agent frameworks increasingly rely on an \emph{agentic loop} in which tools expose explicit lifecycle signals that gate retry and escalation logic. OpenClaw is representative in treating tools as callable actions and structuring their outcomes for decision-making \cite{openclaw}. Closely related paradigms in LLM agents emphasize interleaving reasoning with environment actions and explicit tool invocation to mitigate hallucinations and enable recovery \cite{yao2023react,schick2023toolformer}. We adopt this principle for physical grasping by constructing a \emph{physical tool-state stream} from execution telemetry.

\noindent \textbf{Open-loop language-guided grasping and visual-force primitives.}
Many language-guided grasping systems, and more broadly manipulation systems, map perception and text to action goals and execute primitives in an open-loop or weakly closed-loop fashion. ForceSight predicts Cartesian and force goals from RGB-D and text and achieves strong generalization on a mobile manipulator \cite{collins2024forcesight}, but does not explicitly represent execution outcomes as discrete states or provide a bounded recovery interface. In parallel, language-conditioned manipulation policies and VLAs (spanning imitation learning, behavior cloning, and end-to-end visuomotor control) demonstrate impressive task coverage and generalization, yet typically do not expose a standardized execution-state interface for outcome-driven recovery \cite{shridhar2022cliport,shridhar2023peract,zitkovich2023rt2,driess2023palme}. Recent language-driven grasping approaches that improve target selection (e.g., chain-of-thought style visual grounding) also commonly assume single-shot execution and remain sensitive to distractors and contact uncertainty \cite{zhang2025vcotgrasp}. Our focus is on wrapping a strong primitive with an agentic loop rather than replacing it.

\noindent \textbf{Interactive correction and refinement.}
Human-in-the-loop corrections can improve performance during execution. For example, online language corrections via shared autonomy enable interactive refinement \cite{cui2023noright}. Complementary work treats language as a correction channel that can modify objectives or constraints online, enabling recovery from planning and execution errors without full teleoperation \cite{sharma2022correcting}. Our contribution differs in the signal used to trigger intervention: instead of relying on language feedback as the primary failure signal, we expose execution outcomes as standardized events that can autonomously trigger bounded retries or clarification.

\noindent \textbf{Failure awareness via VLM/VLA supervisors.}
A growing body of work uses vision-language models to detect, diagnose, and correct failures, including VLM-based failure reasoning \cite{duan2024aha} and supervisor-style architectures for prediction/correction \cite{yang2025fpcvla,zeng2025diagnose,grislain2025ifailsense}. More recent efforts emphasize scaling failure data and training failure-aware VLMs for fine-grained detection \cite{liu2023reflect,pacaud2025guardian}. These approaches are often camera-centric and focus on diagnosis and replanning. In contrast, we emphasize a low-latency, telemetry-driven discrete outcome interface designed to serve as the execution-state stream of a physical agentic loop (and to be complementary to vision-based supervisors).

\noindent \textbf{Agentic and closed-loop manipulation frameworks.}
Recent frameworks decompose manipulation into modular agentic systems \cite{yang2025maniagent,yang2025agenticrobot,vorobiov2025arrc} and incorporate verification and re-attempts, including VLM self-verification in real-world settings \cite{duan2024manipulateanything}. Related lines leverage LLMs for skill selection and grounding in long-horizon tasks \cite{ichter2023doasican}, generate executable policy code for reactive control \cite{liang2023codeasp}, compose perception and planning structures for situated task planning \cite{singh2023progprompt}, or perform hierarchical self-correction during task execution \cite{ming2023hicrisp}. VoxPoser additionally demonstrates compositional language grounding into 3D value maps for closed-loop trajectory synthesis \cite{huang2023voxposer}. Our contribution targets a missing abstraction shared across these lines: a deterministic and lightweight mapping from proprioceptive execution telemetry to standardized outcome events, paired with a strictly bounded policy with termination guarantees.

\noindent \textbf{Grasp success/failure verification.}
Grasp verification using multimodal sensing has a long history; recent work includes vision-based verification and sim-to-real transfer \cite{amargant2025sim2real}. Classic and modern slip/outcome detectors show the utility of tactile/visual signals for determining grasp stability \cite{li2018slip}. We use semantic verification as an optional gating signal, but our primary focus is framing outcome monitoring as a tool-state stream that drives bounded retry/escalation logic in an agentic loop.

\smallskip
Taken together, prior work provides strong language-conditioned primitives \cite{collins2024forcesight,zhang2025vcotgrasp}, interactive correction mechanisms \cite{cui2023noright,sharma2022correcting}, and failure supervisors \cite{duan2024aha,yang2025fpcvla,zeng2025diagnose,grislain2025ifailsense,liu2023reflect,pacaud2025guardian} within broader agentic manipulation frameworks \cite{yang2025maniagent,yang2025agenticrobot,vorobiov2025arrc,duan2024manipulateanything,ichter2023doasican,liang2023codeasp,singh2023progprompt,huang2023voxposer,ming2023hicrisp}. To our knowledge, none combine (i) a low-latency telemetry-to-discrete-execution-state abstraction suitable as a physical tool-state stream with (ii) a strictly bounded deterministic retry/clarification policy, packaged as a modular wrapper around an unmodified manipulation primitive for grasping.

%% file: sections/method_v2.tex
\section{Method}
\label{sec:method}

\input{figures/overview}

\subsection{Problem Statement}
We study real-world robotic grasping under perceptual and physical ambiguity.
Given a task specification (e.g., grasp a target object described in language) and an initial grasp proposal from a perception module, the robot must execute the action and autonomously determine whether the task has been successfully completed.

In practical settings, grasp outcomes are often uncertain.
A gripper may fully close without grasping any object, an object may be initially grasped but slip during lifting, or a visually similar distractor may be selected instead of the intended target. Importantly, such failure modes are not always directly observable from pre-grasp vision alone and may not produce explicit failure signals for downstream decision-making.

Most open-loop grasping systems treat execution as a single action: a grasp is planned and executed once, after which the system either terminates or relies on external supervision. These approaches lack a structured mechanism to interpret physical execution outcomes and to decide whether retrying, re-targeting, or requesting clarification is necessary. Inspired by the \emph{agentic loop} paradigm, we instead reformulate language-guided grasping as a bounded autonomous agent operating over grounded execution states, where each physical action exposes an explicit outcome interface for recovery.

\textbf{Agent formulation.}
We define the grasping agent as $\mathcal{A} = (\mathcal{O}, \mathcal{S}, \mathcal{U}, \pi)$, where:
\begin{itemize}
    \item $\mathcal{O}$ denotes observations composed of grasp proposals, execution feedback (telemetry and post-action cues), and optional semantic evidence;
    \item $\mathcal{S}$ denotes discrete execution outcome states emitted by Watchdog (Sec.~\ref{sec:outcome_grounding}), e.g.,
    \texttt{SUCCESS}, \texttt{EMPTY}, \texttt{SLIP}, \texttt{WEAK}, \texttt{STALL}, \texttt{TIMEOUT};
    \item $\mathcal{U}$ denotes high-level decision actions in the \emph{Decide} stage, i.e., $\{\texttt{FINALIZE}, \texttt{RETRY/RESELECT}, \texttt{WAIT\_CLARIFY}\}$;
    \item $\pi : (\mathcal{S}, \mathcal{G}) \rightarrow \mathcal{U}$ denotes a bounded decision policy mapping grounded outcome states and goal specification $\mathcal{G}$ to decision actions.
\end{itemize}
This formulation shifts the focus from improving a specific grasp proposal generator to designing a structured, outcome-grounded grasping agent capable of bounded recovery under ambiguity.

\subsection{Agent Architecture and Event-Based Interface}
\label{sec:agent_arch}

To operationalize the outcome-driven grasping paradigm, we design an agent-centric architecture (Figure~\ref{fig:full_system}) that explicitly separates:
\quad (1) semantic goal reasoning,
\quad (2) perception conditioning, and
\quad (3) physical execution with outcome monitoring.
Rather than treating language and grasping as loosely coupled modules, we represent the system as an \emph{agentic loop} that treats each grasp attempt as a tool call whose outcome is surfaced as a structured event.

\subsubsection{Agent Core}
The Agent Core functions as the central decision-making entity. It parses the user instruction into a structured goal specification $\mathcal{G}$ (e.g., object category and attributes such as color or spatial qualifiers), grounds the goal through perception conditioning, and maintains bounded execution logic over discrete outcome states.

During execution, the Agent Core consumes structured outcome events emitted by the grasping engine (i.e., a \texttt{watchdog\_label} and an \texttt{exec\_status}) and applies a bounded recovery policy. In the default setting, the agent performs at most one automatic retry for recoverable physical failures and otherwise escalates to clarification or terminates safely. When ambiguity persists after a failed attempt, the agent may trigger a bounded re-selection step (e.g., re-evaluating candidate instances and, when applicable, repositioning) before re-execution. This preserves interpretability and prevents unbounded oscillation.

\subsubsection{Event-Based Interface}
To ensure modularity and reproducibility, all inter-module communication follows a structured event schema. Each execution attempt is associated with a unique \texttt{trial\_id}, and the grasping engine publishes an outcome message containing:
\begin{itemize}
    \item \texttt{watchdog\_label} $\in$ \{\texttt{SUCCESS}, \texttt{EMPTY}, \texttt{SLIP}, \texttt{WEAK}, \texttt{STALL}, \texttt{TIMEOUT}\},
    \item \texttt{exec\_status} $\in$ \{\texttt{SUCCESS}, \texttt{FAIL}, \texttt{WAIT\_CLARIFY}\},
    \item timestamp and lightweight debug metadata.
\end{itemize}
These events serve as the \emph{physical tool-state stream}: a standardized execution-state interface that makes physical outcomes directly consumable by the decision policy. System-specific details of stabilization and sensing are deferred to Sec.~\ref{sec:impl_details}. Here \texttt{exec\_status} is a coarse status flag used for logging and downstream handling, while \texttt{watchdog\_label} carries the fine-grained physical outcome used by the policy.

\subsubsection{Outcome-Aware Grasping Engine}
The grasping engine encapsulates a learned motion primitive (e.g., ForceSight~\cite{collins2024forcesight}) together with safety controls and an outcome monitoring layer. The primitive produces grasp targets from RGB-D input; the engine executes the attempt and then emits an outcome event through the interface above. Importantly, we treat the primitive as \emph{unmodified}: the contribution lies in the wrapper that interprets physical execution and exposes it as discrete tool states for decision gating, rather than in changing the underlying grasp predictor.

\subsubsection{Agentic Loop Formulation}
The overall system instantiates a bounded agentic loop:
\begin{enumerate}
    \item \textbf{Observe:} receive structured goal $\mathcal{G}$ and current perception state.
    \item \textbf{Act:} execute the grasping primitive.
    \item \textbf{Evaluate:} infer a discrete execution outcome (Watchdog label) and emit an outcome event.
    \item \textbf{Decide:} select \texttt{FINALIZE}, \texttt{RETRY/RESELECT}, or \texttt{WAIT\_CLARIFY} under a bounded policy.
\end{enumerate}
Unlike classical reactive pipelines, this loop explicitly reasons over execution outcomes: the agent does not merely output actions, but verifies whether the outcome satisfies physically grounded success criteria and conditionally applies bounded recovery.

\subsection{Outcome Grounding as Physical State Abstraction}
\label{sec:outcome_grounding}

A central challenge in outcome-driven grasping is converting noisy, continuous execution feedback into stable, decision-ready states. We formulate this as a \emph{physical state abstraction} problem: map execution evidence to a compact discrete label that is informative for recovery yet stable enough to gate downstream decisions.

Watchdog maps execution feedback into physically grounded outcome labels:
\[
\mathcal{S} =
\{ \texttt{SUCCESS}, \texttt{EMPTY}, \texttt{SLIP}, \texttt{WEAK}, \texttt{STALL}, \texttt{TIMEOUT} \}.
\]
At the agent level, these outcomes are mapped into decision actions:
\[
\mathcal{U} =
\{ \texttt{FINALIZE}, \texttt{RETRY/RESELECT}, \texttt{WAIT\_CLARIFY} \}.
\]
This abstraction decouples continuous control signals from high-level decision logic. Intuitively, \texttt{EMPTY} indicates a recoverable failure candidate; \texttt{SUCCESS} indicates physical completion (subject to semantic consistency when enabled); and the remaining labels capture non-productive or unsafe execution modes under a conservative default policy. The concrete sensing cues and stabilization used to infer these labels are described in Sec.~\ref{sec:impl_details}.

\begin{table}[t]
\caption{Watchdog outcome labels used as discrete execution states (high-level definitions).}
\label{tab:state_definition}
\centering
\small
\renewcommand{\arraystretch}{1.1}
\setlength{\tabcolsep}{4pt}
\begin{tabularx}{\columnwidth}{@{}p{0.25\columnwidth}X@{}}
\toprule
\textbf{Label} & \textbf{Definition (high-level)} \\
\midrule
\textbf{SUCCESS} &
Execution evidence consistent with stable object acquisition (physical completion). \\
\addlinespace
\textbf{EMPTY} &
Execution evidence consistent with no object acquired (recoverable failure candidate). \\
\addlinespace
\textbf{SLIP} &
Transient acquisition followed by loss consistent with slipping. \\
\addlinespace
\textbf{WEAK} &
Marginal or unstable acquisition consistent with a fragile grasp. \\
\addlinespace
\textbf{STALL} &
Execution fails to progress to a stable terminal signature (stalled motion/closure). \\
\addlinespace
\textbf{TIMEOUT} &
Execution exceeds a fixed time budget without reaching a stable terminal signature. \\
\bottomrule
\end{tabularx}
\end{table}

\subsection{Bounded Agentic Decision Policy}
\label{sec:agentic_policy}

After \emph{Evaluate} produces a Watchdog outcome label---and optionally a semantic consistency signal---the agent selects the next decision action under a bounded, deterministic policy. Concretely, the decision stage consumes:
\[
(\, \mathcal{G},\ \texttt{trial\_id},\ \texttt{watchdog\_label},\ \texttt{exec\_status},\ v_{\text{sem}} \,),
\]
where $\mathcal{G}$ is the structured goal and $v_{\text{sem}}$ denotes optional semantic verification.

\paragraph{Decision actions.}
The policy outputs one of:
\[
u \in \{\texttt{FINALIZE},\ \texttt{RETRY/RESELECT},\ \texttt{WAIT\_CLARIFY}\}.
\]
\texttt{FINALIZE} terminates the current trial as completed (or safely terminated),
\texttt{RETRY/RESELECT} triggers a bounded recovery attempt (optionally preceded by re-selection),
and \texttt{WAIT\_CLARIFY} escalates to the user for disambiguation.

\paragraph{Outcome-conditioned policy.}
Given the evaluated outcome, the agent applies the following bounded rules:
\begin{itemize}
    \item \textbf{\texttt{SUCCESS}:} output \texttt{FINALIZE} if semantic evidence is consistent (when enabled); otherwise output \texttt{WAIT\_CLARIFY}.
    \item \textbf{\texttt{EMPTY}:} output \texttt{RETRY/RESELECT} if the retry budget remains; otherwise output \texttt{WAIT\_CLARIFY}.
    \item \textbf{\texttt{SLIP}/\texttt{WEAK}/\texttt{STALL}/\texttt{TIMEOUT}:} output \texttt{FINALIZE} under the conservative default (safe termination), with optional escalation to \texttt{WAIT\_CLARIFY} when the next action depends on user intent (e.g., re-specify target).
\end{itemize}

\paragraph{Boundedness and termination.}
Boundedness is enforced by a fixed retry budget and explicit escalation conditions. Therefore, the system always transitions to either \texttt{FINALIZE} or \texttt{WAIT\_CLARIFY} after a finite number of attempts, preventing unbounded execution in ambiguous or failure-prone scenes.

%% file: figures/overview.tex
\begin{figure*}[t!]
    \centering
    \includegraphics[width=0.8\linewidth]{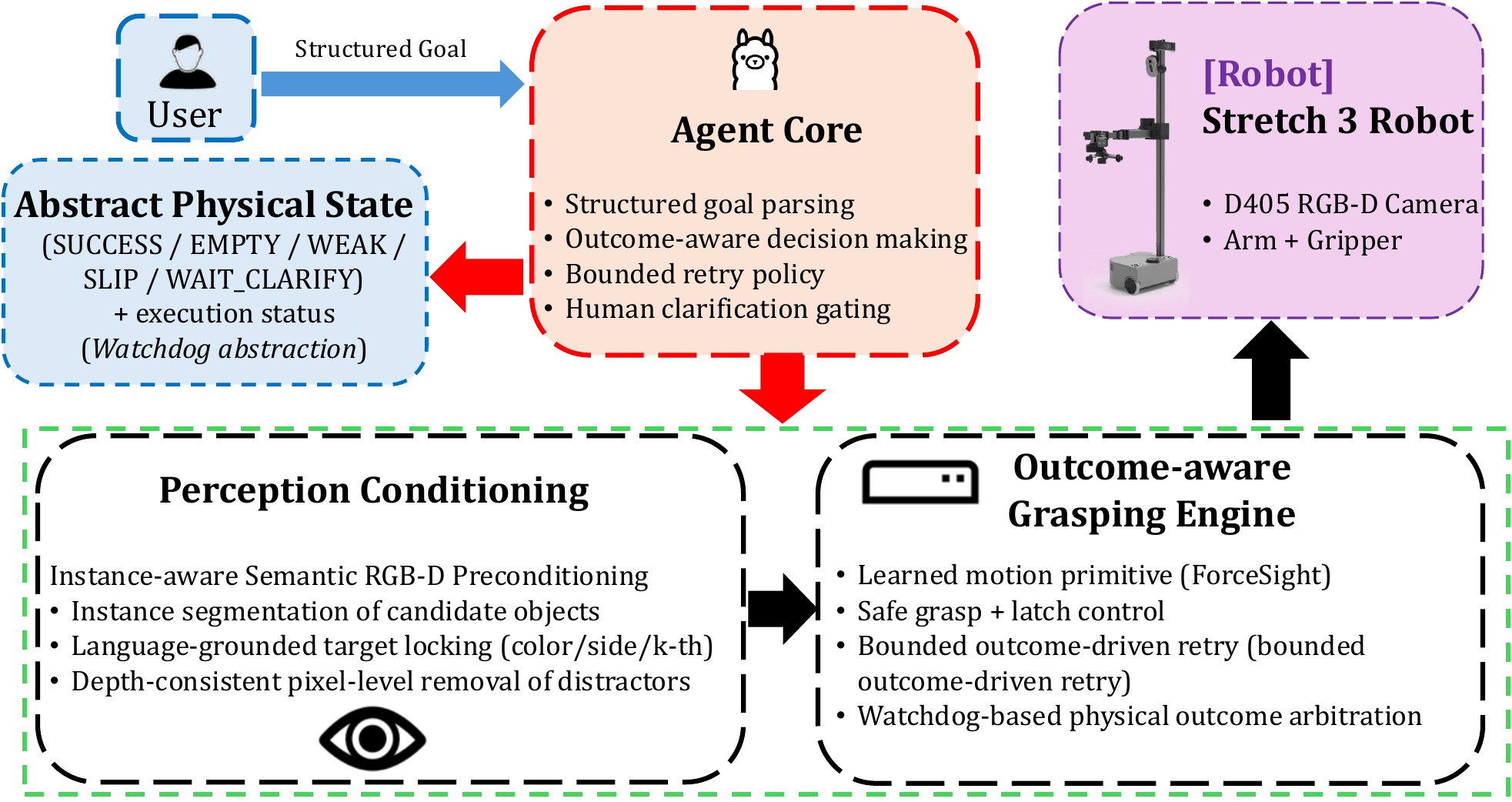}
    \vspace*{-0.5\baselineskip} 
    \caption{\textbf{Agent-centric architecture with a structured event interface.}
    A lightweight language control layer specifies a semantic goal, which the Agent Core parses into structured task constraints $\mathcal{G}$. 
    Perception conditioning produces goal-consistent RGB-D input for the unmodified manipulation primitive, while the outcome-aware execution engine runs the grasp and performs physical outcome monitoring.
    Execution evidence is summarized as a structured event consisting of a Watchdog outcome label (\texttt{SUCCESS}, \texttt{EMPTY}, \texttt{SLIP}, \texttt{WEAK}, \texttt{STALL}, \texttt{TIMEOUT}) and an execution status (\texttt{SUCCESS}, \texttt{FAIL}, \texttt{WAIT\_CLARIFY}). 
    The Agent Core consumes these events to realize the bounded agentic loop (observe--act--evaluate--decide), selecting to finalize, retry/reselect, or request clarification/terminate.}
        \label{fig:full_system}
\end{figure*}

%% file: sections/experiments_v2.tex
\section{Experiments}
\label{sec:experiments}

\subsection{Experimental Setup}
\label{sec:exp_setup}

All experiments were conducted on a Hello Robot Stretch mobile manipulator \cite{stretch3} equipped with an eye-in-hand Intel RealSense D405 RGB-D camera. The underlying manipulation primitive is ForceSight \cite{collins2024forcesight}, used \emph{without} retraining or fine-tuning. The \emph{baseline} is open-loop execution: a single grasp attempt with no structured outcome monitoring and no recovery logic. \emph{Ours} wraps the same primitive with Watchdog outcome events and a bounded decision policy; we additionally evaluate an optional semantic verification variant where stated.

\subsection{Metrics and Protocol}
\label{sec:metrics}

We report \emph{task success rate} (correct target grasped and lifted). For each scenario, we run $N=10$ trials per method under matched scene configurations. For diagnostic analyses, we also report (when applicable) module-level success counts and runtime statistics.

In total, we conducted more than 290 real-robot trials over the course of system development and evaluation, of which 180 were performed with the final system configuration. The quantitative results reported in this paper are drawn exclusively from these final-system trials under the controlled benchmark scenarios described above.

\subsection{Scenarios}
\label{sec:scenarios}

\begin{figure}[t]
    \centering
    \includegraphics[width=\linewidth]{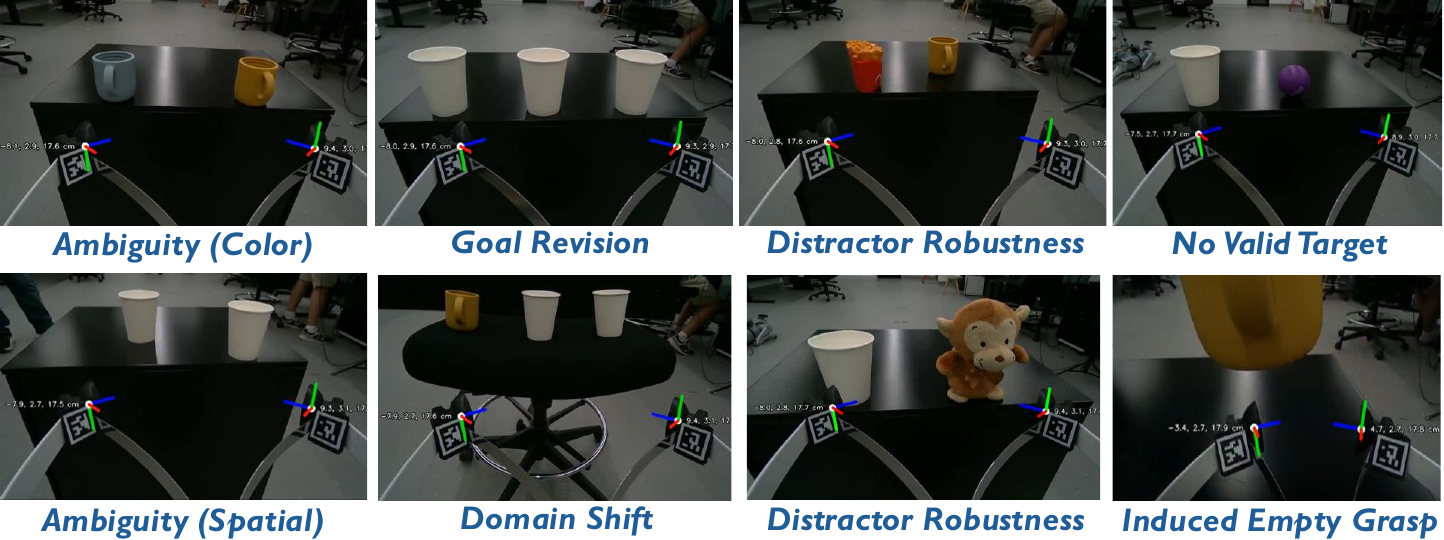}
    \caption{
    \textbf{Benchmark scene snapshots used in our real-robot evaluation.} Each panel corresponds to a representative scenario from Sec.~\ref{sec:scenarios}, including target ambiguity (color/spatial), distractor interference, domain shift (platform/background/lighting changes), induced empty-grasp failures for recovery evaluation, and infeasible targets (no valid object of interest) to test safe clarification/termination behavior.
    }
    \label{fig:scenarios}
\end{figure}

We evaluate representative scenarios that directly probe the agentic-loop capabilities. Figure~\ref{fig:scenarios} shows snapshots of the physical scene configurations used for these benchmarks.
\begin{itemize}
    \item \textbf{Ambiguity (Color):} multiple similar objects; instruction specifies color.
    \item \textbf{Ambiguity (Spatial):} identical objects; instruction specifies spatial qualifiers (left/right/front/back).
    \item \textbf{Distractor Robustness:} salient non-target object near the target.
    \item \textbf{Domain Shift:} different platform/background/lighting (no retraining).
    \item \textbf{Induced Empty Grasp:} perturb execution to trigger \texttt{EMPTY}; test bounded retry.
    \item \textbf{No Valid Target:} target category absent; test safe clarification/termination.
    \item \textbf{Goal Revision:} instruction changes mid-execution; test event-driven re-targeting.
\end{itemize}

\begin{table}[t]
\caption{Quantitative results across representative scenarios (10 trials per configuration). Success rate is reported as the percentage of successful grasps.}
\label{tab:quant}
\centering
\footnotesize
\setlength{\tabcolsep}{3.0pt}
\renewcommand{\arraystretch}{1.15}
\begin{tabular}{lcc}
\toprule
\textbf{Scenario} & \textbf{Baseline} & \textbf{Ours (Agentic loop)} \\
\midrule
Single Target                    & 80\% (8/10)  & 100\% (10/10) \\
Ambiguity (Color / Spatial)      & 40\% (4/10)  & 80\% (8/10)   \\
Distractor Robustness            & 0\% (0/10)   & 100\% (10/10) \\
Multiple Identical Targets       & 10\% (1/10)  & 100\% (10/10) \\
\bottomrule
\end{tabular}
\vspace{-1.2ex}
\end{table}

\noindent
Table~\ref{tab:quant} shows that open-loop execution performs reasonably under trivial single-target conditions, but degrades substantially under ambiguity and distractor interference. In contrast, the proposed agentic loop maintains high success rates across the structured scenarios, consistent with the benefits of explicit execution-state monitoring and bounded recovery. Across successful trials of our method, recovery was achieved with at most one automatic retry per trial, confirming the bounded nature of the policy.

\begin{table}[t]
\caption{Component ablation across ambiguity scenarios. Results are reported as successful grasps over total trials.}
\label{tab:ablation}
\centering
\footnotesize
\setlength{\tabcolsep}{3pt}
\renewcommand{\arraystretch}{1.15}
\begin{tabular}{lccc}
\toprule
\textbf{Configuration} & \textbf{Diff-color} & \textbf{Same-color} & \textbf{Total} \\
\midrule
No Watchdog            & 3/5 & 4/5 & 7/10 \\
No Vision Conditioning & 1/5 & 3/5 & 4/10 \\
No Retry               & 5/5 & 4/5 & 9/10 \\
Full System            & 5/5 & 5/5 & 10/10 \\
\bottomrule
\end{tabular}
\vspace{-1ex}
\end{table}

\noindent
\textbf{Component ablation.}
Table~\ref{tab:ablation} isolates the contribution of key components under ambiguity. Removing Watchdog or semantic perception conditioning substantially reduces success, indicating that both execution-state monitoring and goal-grounded perception are important for robust grasping in cluttered scenes. Disabling retry also reduces performance, confirming that bounded recovery contributes beyond improved target selection alone. The full system achieves the highest success rate across both ambiguity settings.

\begin{figure}[t]
\centering
\includegraphics[width=0.8\linewidth]{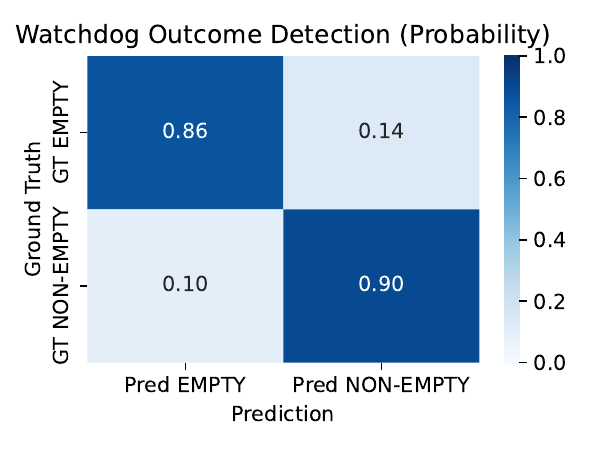}
\caption{
\textbf{Confusion matrix for Watchdog predictions.}
Rows represent ground-truth execution outcomes,
and columns represent watchdog predictions. 
}
\label{fig:watchdog_confusion}
\end{figure}

\noindent
\textbf{Watchdog outcome accuracy.}
We evaluate the reliability of the watchdog outcome monitoring layer using controlled execution outcomes.
For empty-grasp cases, the watchdog correctly detected
\texttt{EMPTY} in 43 out of 50 trials.
For successful grasps, 5 out of 50 trials were incorrectly
classified as \texttt{EMPTY}.
Figure~\ref{fig:watchdog_confusion} shows the confusion matrix summarizing the detection results.

\begin{table}[t]
\caption{Runtime comparison between the baseline and the proposed agentic-loop system (single-target scenario, $N=10$).}
\label{tab:runtime}
\centering
\footnotesize
\setlength{\tabcolsep}{6pt}
\renewcommand{\arraystretch}{1.15}
\begin{tabular}{lcc}
\toprule
\textbf{Method} & \textbf{Mean Runtime (s)} & \textbf{Std (s)} \\
\midrule
Baseline (open-loop) & 14.78 & 2.97 \\
Ours (Agentic loop)  & 15.94 & 7.27 \\
\bottomrule
\end{tabular}
\vspace{-1ex}
\end{table}

\noindent
\textbf{Runtime overhead.}
Table~\ref{tab:runtime} reports end-to-end runtime for a representative single-target setting. Despite adding execution monitoring and decision logic, the overall runtime remains comparable to the baseline. This supports the claim that the agentic-loop wrapper introduces minimal overhead relative to the grasp execution cycle.

\subsection{Qualitative Results and Behavioral Traces}
\label{sec:qual}

\begin{figure*}[t]
    \centering
    \includegraphics[width=\linewidth]{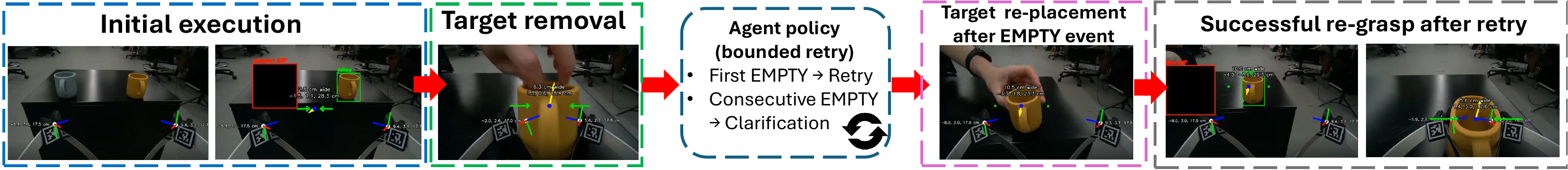}
    \caption{
    \textbf{Outcome-driven recovery timeline}. An induced empty grasp is classified as \texttt{EMPTY} and triggers a bounded retry. If empties persist, the agent escalates to clarification, guaranteeing termination.
    }
    \label{fig:recovery_timeline}
\end{figure*}

\begin{figure*}[t]
    \centering
    \includegraphics[width=\linewidth]{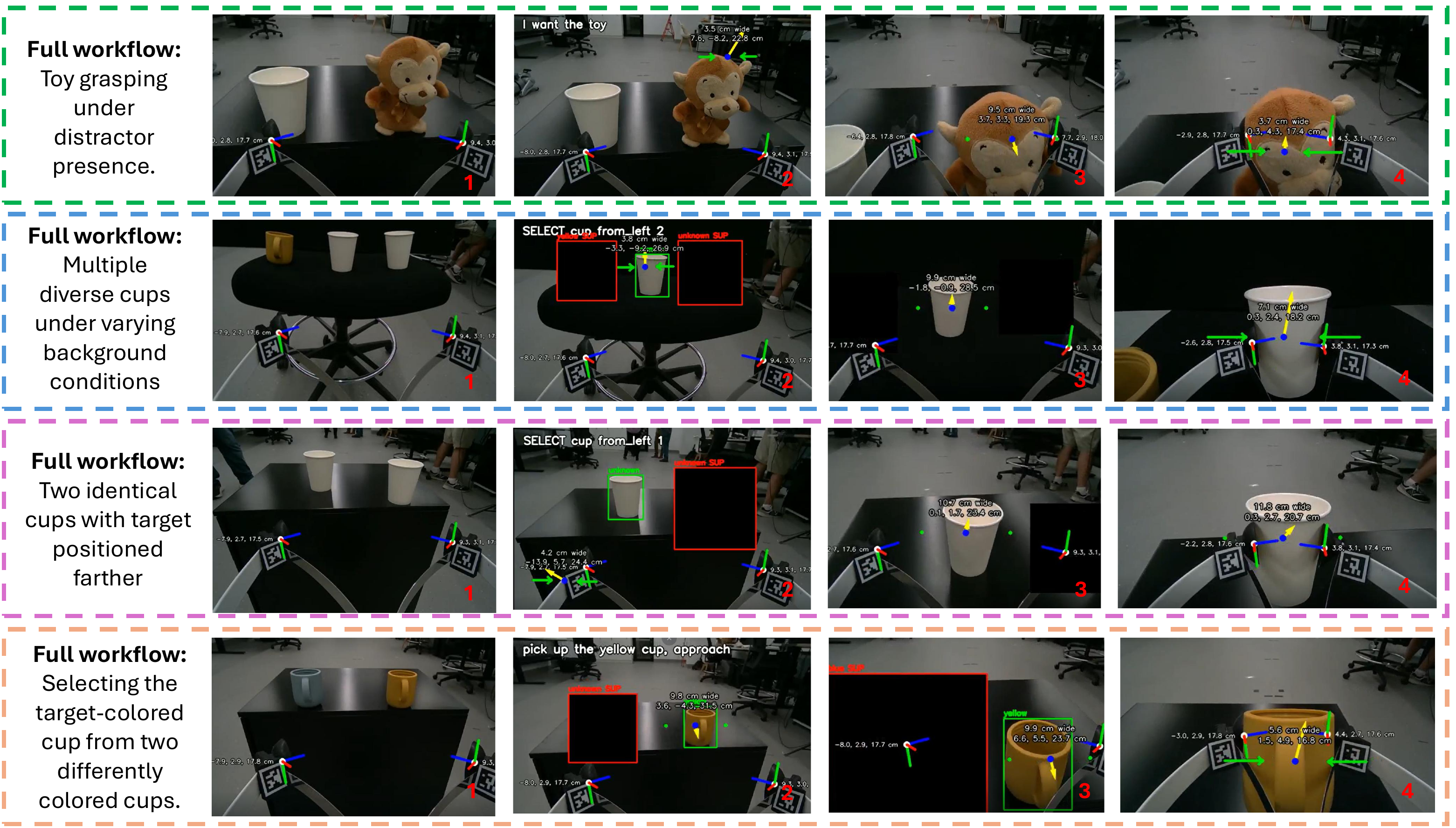}
    \caption{
    \textbf{
    End-to-end execution workflow of the proposed system across representative task scenarios.}
    Each row illustrates a complete execution trace, and each column corresponds to a stage of the agentic grasping pipeline.
    \emph{Column 1: Scene observation.}
    The robot observes the scene and receives a natural-language instruction specifying the target object or constraint.
    \emph{Column 2: Target grounding and perception conditioning.}
    The system parses the instruction and applies perception conditioning to isolate the intended target. 
    Depending on the task, non-target objects may be suppressed (black regions) using RGB-D conditioning or instance selection.
    \emph{Column 3: Grasping execution.}
    The robot executes the visual servoing policy to approach and grasp the selected target object while maintaining target consistency.
    \emph{Column 4: Outcome evaluation and final grasp state.}
    The watchdog monitors gripper signals and system state to determine the outcome, enabling bounded decisions such as finalize, retry, or clarification if needed.
    The four rows correspond to different task scenarios:
    (top) toy grasping under distractor presence,
    (second) multiple diverse cups under varying background conditions,
    (third) two identical cups with the target positioned farther away,
    and (bottom) selecting the target-colored cup from two differently colored cups.
    Across scenarios, the system maintains consistent target selection through semantic grounding and outcome-aware execution.
    }
    \label{fig:full_workflow}
\end{figure*}

Figure~\ref{fig:recovery_timeline} illustrates a representative recovery episode: Watchdog detects an induced empty grasp, triggers a bounded retry, and escalates to clarification when empties persist, making termination behavior explicit. Figure~\ref{fig:full_workflow} shows an end-to-end trace of the observe--act--evaluate--decide pipeline.

\subsection{Implementation Details}
\label{sec:impl_details}

The system-specific details are as follows:
\begin{itemize}
    \item \textbf{Perception conditioning:} depending on instruction and clutter, the agent selects between lightweight color filtering and instance segmentation for suppressing non-target regions.
    \item \textbf{Large Language Model:} A lightweight local language model (TinyLlama-1.1B~\cite{zhang2024tinyllama}) is used for natural-language command parsing and interaction management.
    \item \textbf{Watchdog stabilization:} outcome emission requires a short settle window where effort/closure cues remain consistent; micro-lift evidence is used to reject contact-only false positives.
    \item \textbf{Budgets and timeouts:} one automatic retry for \texttt{EMPTY} by default; fixed per-attempt timeout; explicit transitions to clarification to prevent oscillations.
\end{itemize}

%% file: sections/conclusion.tex
\section{Discussion and Limitations}
\label{sec:discussion}

This work does not introduce a new grasp predictor or train additional models. Instead, it contributes a system-level abstraction that makes physical execution outcomes legible as a tool-state stream and uses those states to drive a bounded agentic loop. The experiments suggest that a significant portion of failures in language-guided grasping arise from missing runtime outcome structure rather than inadequate grasp prediction.

Our current policy is intentionally conservative: \texttt{EMPTY} triggers at most one automatic retry, while other failure labels are treated as terminal outcomes under the default setting. This prioritizes safety, interpretability, and guaranteed termination over aggressive recovery. A natural extension is to introduce label-specific recovery (e.g., regrasp after \texttt{SLIP}, re-approach after \texttt{WEAK}) while maintaining boundedness.

Limitations include: (i) semantic disambiguation depends on upstream perception reliability, (ii) Watchdog currently emphasizes robust separation between empty and non-empty grasps, and more fine-grained strategies may further improve performance, and (iii) feasibility reasoning beyond presence/category constraints remains limited. Despite these limitations, the results indicate that structuring execution as a bounded physical agentic loop can substantially improve robustness and interpretability without modifying the underlying grasp model.

\section{Conclusion}
\label{sec:conclusion}
We presented a physical agentic loop for language-guided grasping that treats physical actions as tool calls with explicit execution-state events, enabling reliable closed-loop behavior without retraining the underlying visual-force grasping model. Our key component, \emph{Watchdog}, converts continuous gripper telemetry into discrete outcome labels with confidence and supports a bounded policy that deterministically decides to finalize, retry once, or request clarification, thereby guaranteeing termination and improving interpretability. Experiments on a Stretch platform show that feeding explicit execution outcomes back into the agent substantially improves robustness under ambiguity and distractors compared to open-loop execution, while maintaining safety through strict retry limits. Overall, the results suggest that a missing ingredient in many language-guided grasping systems is not only stronger perception or control, but a standardized execution-state interface that makes physical outcomes observable and actionable at runtime. Future work will extend the framework with outcome-specific recovery strategies and stronger semantic verification while preserving bounded, event-driven loop semantics.
